\documentclass{article}
\pdfoutput=1

\usepackage[numbers]{natbib}

\usepackage[utf8]{inputenc} 
\usepackage[T1]{fontenc}    
\usepackage{hyperref}       
\usepackage{url}            
\usepackage{booktabs}       
\usepackage{amsfonts}       
\usepackage{nicefrac}       
\usepackage{microtype}      

\usepackage{bm}
\usepackage{comment}
\usepackage{amsmath} 
\usepackage{subcaption}
\usepackage{listings} 
\usepackage{graphicx}
\usepackage{caption}

\begin{document}

\title{Synth-Validation: Selecting the Best Causal Inference Method for a Given Dataset}

\author{Alejandro Schuler \\
	aschuler@stanford.edu \\
       Biomedical Informatics Research\\
       Stanford University\\
       Palo Alto, CA, USA 
       \and
     Ken Jung\\
	 kjung@stanford.edu \\
       Biomedical Informatics Research\\
       Stanford University\\
       Palo Alto, CA, USA 
       \and
     Robert Tibshirani \\
     tibs@stanford.edu \\
       Department of Statistics\\
       Stanford University\\
       Palo Alto, CA, USA 
          \and
     Trevor Hastie \\
     hastie@stanford.edu \\
       Department of Statistics\\
       Stanford University\\
       Palo Alto, CA, USA 
          \and
     Nigam Shah \\
     nigam@stanford.edu \\
       Biomedical Informatics Research\\
       Stanford University\\
       Palo Alto, CA, USA }

\maketitle

\begin{abstract}

Many decisions in healthcare, business, and other policy domains are made without the support of rigorous evidence due to the cost and complexity of performing randomized experiments. Using observational data to answer causal questions is risky: subjects who receive different treatments also differ in other ways that affect outcomes. Many causal inference methods have been developed to mitigate these biases. However, there is no way to know which method might produce the best estimate of a treatment effect in a given study. In analogy to cross-validation, which estimates the prediction error of predictive models applied to a given dataset, we propose \emph{synth-validation}, a procedure that estimates the estimation error of causal inference methods applied to a given dataset. In synth-validation, we use the observed data to estimate generative distributions with known treatment effects. We apply each causal inference method to datasets sampled from these distributions and compare the effect estimates with the known effects to estimate error. Using simulations, we show that using synth-validation to select a causal inference method for each study lowers the expected estimation error relative to consistently using any single method.

\end{abstract} 
 
\section{Introduction}

\subsection{Background}
 
The fundamental problem of causal inference is that after a subject receives a treatment and experiences an outcome it is impossible to know what the outcome would have been had the subject received a different treatment (the counterfactual outcome) \cite{Stuart:2013dt}. The difference in outcome between those two treatments is the true causal effect of the treatment on that subject. It is possible to avoid this problem by randomizing which subjects receive the treatment. Randomization ensures that the populations receiving the different treatments are statistically identical and that the difference in average outcomes between the two populations approaches the true average causal effect in expectation \cite{ROSENBAUM1983}.

Subjects in an observational study are not randomized to their treatments, which creates \textsl{confounding}: the resulting treatment populations are different at baseline in ways that affect their corresponding outcomes \cite{King2005}. Consequently, differences in outcomes cannot be solely attributed to differences in treatment. It is well known that observational data analysis is a risky business \cite{Rubin2010, Hannan:2008gh}. 

Causal inference methods exist to alleviate the problems with observational data analysis \cite{Stuart:2013dt}. Many of these methods work by matching together subjects similar at baseline from each treatment population and using only the matched subsample for further analysis. Popular causal inference methods include covariate matching \cite{Iacus}, propensity score matching \cite{ROSENBAUM1983}, and inverse probability weighting \cite{Robins1992}, while less widely-used methods include doubly robust estimators \cite{Robins1992}, prognostic score matching \cite{Leacy:2013fs}, and targeted maximum likelihood estimation \cite{Schuler:2017cq}. Each method has a number of variations and many methods can be combined and used in tandem with other methods \cite{Colson:2016fu}. 

\begin{figure}[h!]
\centering
\includegraphics[width=0.75\textwidth]{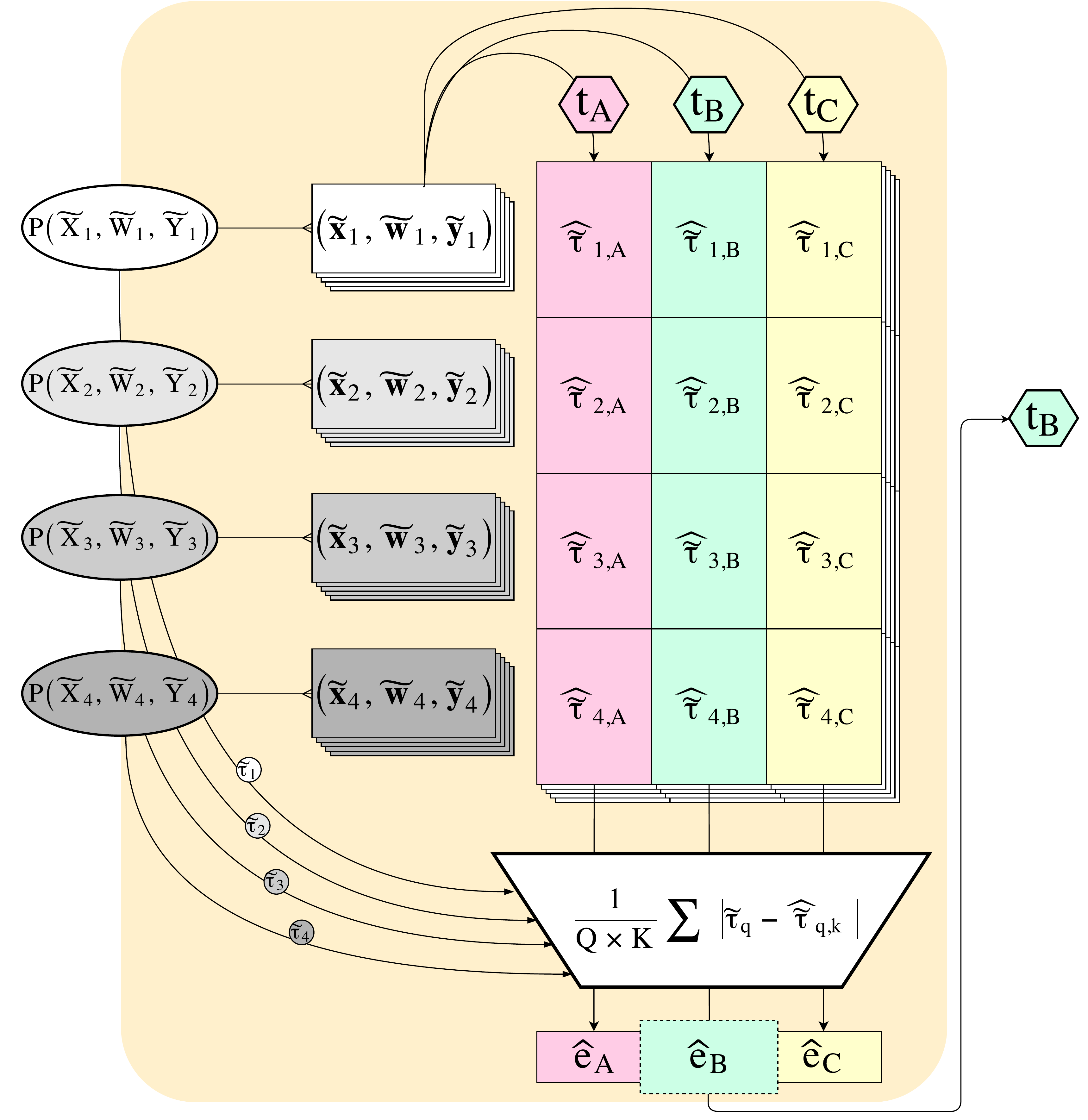} 
\caption{A standard evaluation of some causal inference methods $t_A$, $t_B$, $t_C$ using handcrafted generative distributions. The experimenter comes up with several data-generating distributions from which several datasets are sampled. All datasets are fed through each causal inference method, producing treatment effect estimates. Estimates are compared to the corresponding true effects and the errors are averaged over all samples and data-generating distributions. The causal inference method with lowest average error is deemed best.}
\label{fig:bakeoff}
\end{figure}

Existing comparative evaluations of causal inference methods rely on handcrafted data-generating distributions that encode specified treatment effects \cite{Setoguchi2008, Colson:2016fu, Shortreed:2017fk, Antonelli:2016ve, Lee2010, Hill2011}. Causal inference methods are applied to datasets sampled from these distributions and the results are compared with the known effects to see how close each method gets on average (figure \ref{fig:bakeoff}). However, dissimilarities between these handcrafted benchmarks and real-world data-generating processes mean that it is difficult to know what methods actually work best with real data in general or in the context of a specific study. Because different causal inference methods rely on assumptions that hold in different cases, is likely that different methods are better suited for different studies. There is little consensus among evaluations that use different handcrafted benchmarks, which suggests that there is no one-size-fits-all best method \cite{Setoguchi2008, Colson:2016fu, Shortreed:2017fk, Antonelli:2016ve, Lee2010, Hill2011}. Deep domain and statistical expertise is necessary to pick an appropriate method, which means that many researchers default to what they learned in their first course in statistics (i.e. linear regression) \cite{Stuart2010}. A data-driven approach could augment expertise and safeguard against na\"{\i}vet\'{e}. 

This problem does not exist in a predictive modeling setting, where the goal is to estimate the outcomes for previously-unseen subjects. Predictive modelers can easily ``synthesize'' unseen subjects by hiding a set of subjects (the test set) from their models during fitting. Models are benchmarked by comparing predictions with the previously hidden ground-truth outcomes in the test set to compute a test error. This is most often done in a round-robin fashion (cross-validation). The cross-validation error is an estimate of how well the model predicts outcomes generated by the unknown distribution that the training data is sampled from \cite{Hastie:2009fg}. Depending on the underlying distribution, different models work better, which is reflected in their relative cross-validation errors. The end result is that, given a dataset, predictive modelers are able to determine which model is likely to produce the most accurate prediction of future outcomes.

The same cannot be done for causal inference methods because the estimand (the treatment effect) is a parameter of the entire data-generating distribution, not a variable that can be directly measured \cite{Shmueli:2010ec}. For that reason, it does not suffice to hide a portion of the dataset. Estimating error requires us to know the true treatment effect a-priori which is a catch-22 since the treatment effect is what we would like to estimate in the first place. Because of this, there is no way to get around simulating data from known distributions with known effects. However, we show how the observed data can inform the creation of these distributions to produce study-specific benchmarks. The end result is that, given a dataset, we are able to determine which causal inference method is likely to produce the most accurate estimate of the treatment effect. We call our procedure synth-validation.

The difference between synth-validation and the handcrafted benchmarking process we describe above (figure \ref{fig:bakeoff}) is that the generative distributions we use in synth-validation are derived from the observed data. Our hypothesis, which we confirm experimentally, is that letting the study data shape the generative distributions improves our ability to find the right causal inference method for that study.

Cross-validation and synth-validation are related only insofar as they allow users to compare models or estimators in the context of their own data. Unlike cross-validation, synth-validation is not entirely ``model free'': the process involves modeling generative distributions. The modeling choices we make could bias the procedure towards favoring certain causal inference methods, but we find that these biases are empirically negligible when we use flexible models. 

Synth-validation is inspired by and generalizes the ``plasmode simulations'' of  \citet{Franklin:2014kz}.

\subsection{Outline}

Section \ref{sec:notation} briefly describes some notation. In section \ref{sec:algo} we derive a class of algorithms that use observed data to estimate generative distributions encoding specified treatment effects. Each algorithm is defined by three parts: an algorithm to select what treatment effects to encode, an algorithm that fits a predictive outcomes model that is constrained to encode a specified effect, and a semi-parametric bootstrap that combines a predictive model with a noise model and distribution over predictor variables to create a fully generative model.

To test the efficacy of synth-validation, we use several simulated datasets that we treat as ``real'' for the purposes of the evaluation (section \ref{sec:eval}). We use synth-validation to select the best causal inference method to use for each dataset and use that method to estimate the treatment effect from the real data, which we compare with the known ground-truth effect to calculate the error. We average the errors across all datasets and compare with the average errors obtained by consistently using each single causal inference method.

In section \ref{sec:results} we show the result of the evaluation. Using synth-validation to select different causal inference methods for each dataset is better than using the single best causal inference method for all datasets. In section \ref{sec:disc} we discuss the result, performance differences between variants of synth-validation, limitations, and future directions. 
\section{Notation}
\label{sec:notation}

Let bold functions $\mathbf{f}$ represent the row-wise vectorized versions of their scalar counterparts $f$: $\mathbf{f}(\mathbf{x}) = [f(x_1), f(x_2) \dots f(x_n)]^T$. 

Let $I_a(z)$ be the indicator function for $z=a$.

Let $P(Z^*)$ be a distribution that places probability $\frac{1}{n}$ on each element of  a set of samples $\{z_1, z_2 \dots z_n\}$. Let $z^*$ denote a sample from $Z^*$ and $\bm{z}^*$ denote a set of $n$ such samples (a bootstrap sample).

\subsection{Notation for Observational data}

Observational datasets are represented by a set of $n$ tuples of observations $(x_i,w_i,y_i)$. For each subject $i$, $x_i \in \mathcal{R}^{p}$ is a vector of observed pre-treatment covariates, $w_i \in \{0,1\}^n$ is a binary indicator of treatment status, and $y_i \in \mathcal{R}^n$ is a real-valued outcome. Let capital letters $X$, $W$, and $Y$ represent the corresponding random variables. Let $\mathcal{S}_0$ be the set of indices of untreated subjects, and $\mathcal{S}_1$ the set of indices of treated subjects: $\mathcal{S}_w = \{i | w_i = w\}$.  Denote a series of I.I.D. realizations of a random variable $z \sim P(Z)$ as $\bm{z}$ so that the full observational dataset can be written as $d = (\mathbf{x}, \mathbf{w}, \mathbf{y})$ where $(x_i,w_i,y_i) \overset{\text{I.I.D.}}{\sim} P(X,W,Y)$. 

The average treatment effect is defined as 

\begin{equation}
\tau = E_{X,Y}[Y|X,W=1] - E_{X,Y}[Y|X,W=0]
\label{eq:effect}
\end{equation}

which is the expected difference between what the outcome would have been had a subject received the treatment and what the outcome would have been had a subject not received the treatment, averaged over all subjects in the population. 
\section{Synth-Validation}
\label{sec:algo}

Synth-validation consists of two steps: 1. creating generative distributions that are informed by the observed data (section \ref{sec:gen-model}) and 2. using data sampled from those distributions to benchmark causal inference methods (section \ref{sec:run-meth}). 

\begin{figure}[h!]
\centering
\includegraphics[width=\textwidth]{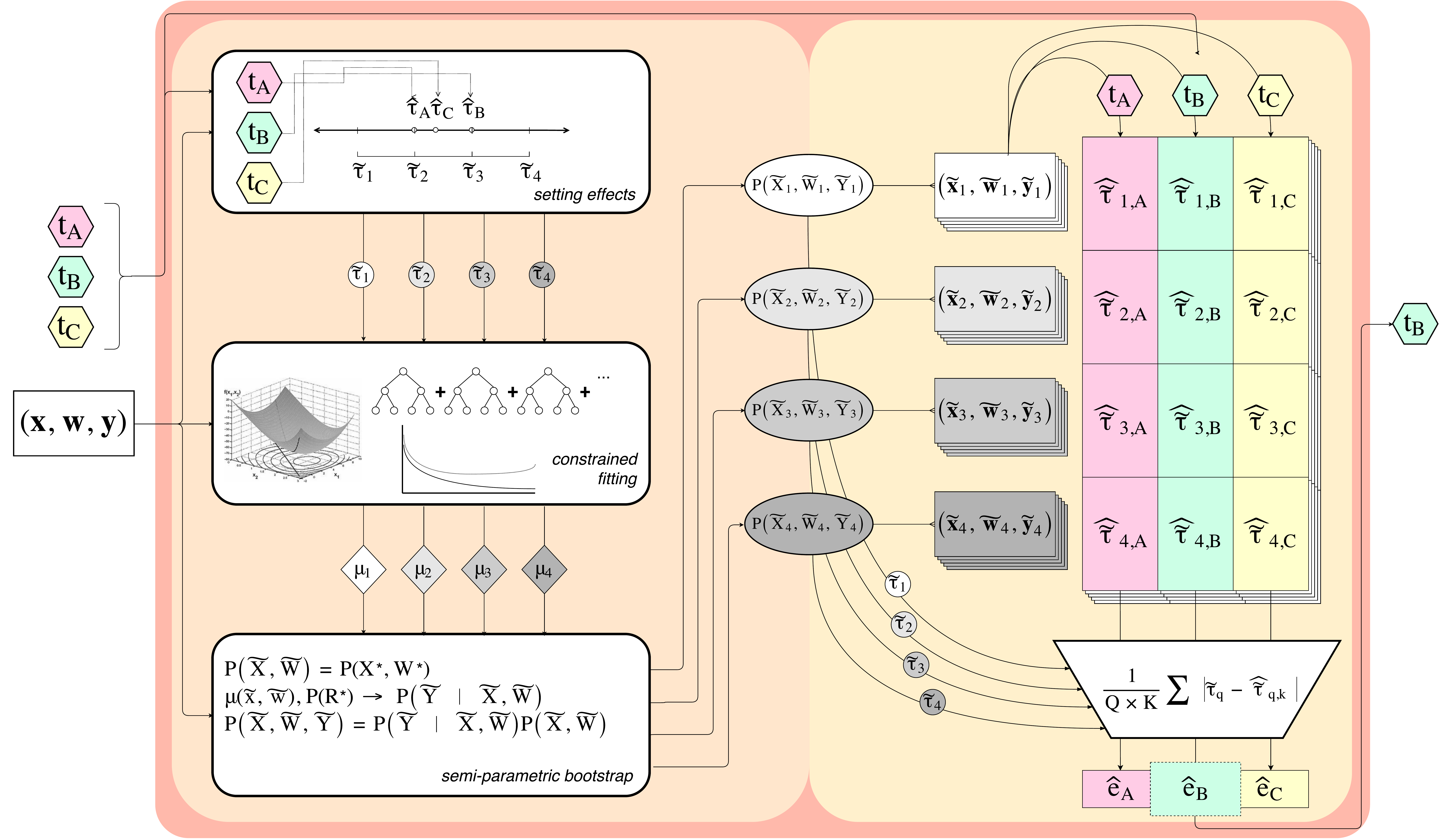} 
\caption{A visual representation of synth-validation. We begin with observed data and several causal inference methods to evaluate. We use the observed data and causal inference methods to create a set of synthetic effects. For each synthetic effect, we use a constrained fitting algorithm to estimate a conditional mean model. We use a semi-parametric bootstrap with each conditional mean model to create a generative distribution and sample data from it. The sampled data, along with the known synthetic average treatment effects, are used to benchmark the causal inference methods. The method with the lowest average error is selected.}
\label{fig:synth-val}
\end{figure}

The entire process is visualized in figure \ref{fig:synth-val}

\subsection{Generative Modeling} 
\label{sec:gen-model}

In this section, we describe the algorithms we use to create the generative distributions for synth-validation. Our goal is not to directly estimate the unknown true data-generating distribution $P(X,W,Y)$. However, we will loosely use the term ``estimate'' to describe the process of creating our generative distributions, since they are informed by the observed data. If we were confident that we could estimate $P(X,W,Y)$ with a particular method, then causal inference could proceed by directly estimating $P(X,W,Y)$ and calculating the treatment effect under that model using equation \ref{eq:effect}.

Instead, we \emph{specify} a desired treatment effect prior to estimating each generative model. We call these ``synthetic treatment effects'' and write them as $\tilde{\tau}_i$. We discuss how to decide good values for these synthetic effects in section \ref{sec:synth-effect}. For each synthetic effect, we find a generative distribution $P(\tilde X, \tilde W, \tilde Y)$ that both satisfies the synthetic effect (i.e. $\tilde\tau = E_{\tilde X,\tilde Y}[\tilde Y| \tilde X, \tilde W=1] - E_{\tilde X,\tilde Y}[\tilde Y| \tilde X, \tilde W=0]$) and maximizes the likelihood of the observed data. The result is a set of distributions $\{P(\tilde X_1, \tilde W_1, \tilde Y_1), P(\tilde X_2, \tilde W_2, \tilde Y_2) \dots\}$. Using multiple distributions is analogous to regularization. Using a single estimate of $P(X,W,Y)$ for benchmarking purposes could ``over-fit'' to the observed data, whereas using a variety of handcrafted generative models that are not tailored to the observed data would ``under-fit'' by attempting to find a one-size-fits-all best method. Setting the treatment effect and allowing the data to dictate the rest of the distribution allows us to vary the generative distributions along the single parameter that we are interested in estimating. The choice of synthetic effects impacts how well the distributions $P(\tilde X, \tilde W, \tilde Y)$ can stand in for $P(X, W, Y)$ in the benchmarking process (figure \ref{fig:bakeoff}), but we show empirically that the heuristic we develop in section \ref{sec:synth-effect} makes synth-validation robust against misspecification.

To simplify the problem of optimally estimating distributions $P(\tilde X, \tilde W, \tilde Y)$ from the observed data $(\mathbf{x}, \mathbf{w}, \mathbf{y})$ we factor the distribution $P(\tilde X, \tilde W, \tilde Y) = P(\tilde Y | \tilde X, \tilde W)P(\tilde X, \tilde W)$ and estimate each factor separately. We estimate $P(\tilde X, \tilde W)$ with $P(X^*, W^*)$, which is the empirical distribution of $(X,W)$ and the nonparametric MLE \cite{Efron:1993dc}. This allows us to simplify the expression for the treatment effect (equation \ref{eq:effect}) by replacing the expectation over $\tilde X$ with a discrete sum:

\begin{equation}
\begin{array}{rcl}
\tilde \tau & = & E_{\tilde X, \tilde Y}[ \tilde Y| \tilde X, \tilde W=1]  - E_{\tilde X,\tilde Y}[\tilde Y| \tilde X, \tilde W=0] \\
& = & \frac{1}{n}\sum_iE_{\tilde Y}[ \tilde Y| \tilde X=x_i, \tilde W=1]  - \frac{1}{n}\sum_i E_{\tilde Y}[\tilde Y| \tilde X=x_i, \tilde W=0] \\
& = & \frac{1}{n}\sum_i[\mu_1(x_i) - \mu_0(x_i)]
\end{array}
\label{eq:constraint}
\end{equation}

where $\mu_0(\tilde x) = E[\tilde{Y}|\tilde X, \tilde W=0]$ and $\mu_1(\tilde x)=E[\tilde{Y}|\tilde X, \tilde W=0]$ are the conditional mean functions of the outcome given each treatment condition. 

The task is now to estimate $P(\tilde Y | \tilde X, \tilde W)$. For this we assume a model: 

\begin{equation}
\begin{array}{rcl}
\tilde y_i & = & I_0(\tilde w_i)\mu_0(\tilde x_i) + I_1(\tilde w_i)\mu_1(\tilde x_i)+ \varepsilon_i \\
\varepsilon_i & \overset{\text{I.I.D.}}{\sim} & \mathcal{E}
\end{array}
\end{equation}

We will soon discuss how to estimate $\mu_0(\tilde x)$ and $\mu_0(\tilde x)$, but for the time being, assume we have estimated these functions. We calculate the observed residuals $r_i = y_i - I_0(w_i)\mu_0(x_i) + I_1(w_i)\mu_1(x_i)$ and use their empirical distribution $P(R^*)$ as an estimate of $\mathcal{E}$, which is a noise model for the outcomes.

The final task is to estimate $\mu_0(\tilde x)$ and $\mu_1(\tilde x)$. Standard machine learning approaches will not work because we must also satisfy the constraint we derived in equation \ref{eq:constraint}. Recall that we are \emph{specifying} the synthetic average treatment effect $\tilde \tau$ for each generative distribution. However, there is no need to ignore the observed data: we can estimate $\mu_0(\tilde x)$ and $\mu_1(\tilde x)$ by constraining the fitting of predictive models for each function. 

\begin{equation}
\mu_0(\tilde x), \mu_1(\tilde x) = 
\begin{cases}
\underset{f_0, f_1 \in \mathcal{F}}{\text{argmin}} \ \  \sum_{\mathcal{S}_0}  l(y_i, f_0(x_i)) + \sum_{\mathcal{S}_1}  l(y_i, f_1(x_i)) \\
\text{subject to:   } \frac{1}{n}\sum_i[f_1(x_i) - f_0(x_i)] = \tilde \tau
\end{cases}
\label{eq:opt}
\end{equation}

$\mathcal{F}$ is a set of functions, or model space, over which the algorithm searches. $l(y,f)$ is a loss function that defines the quality of fit for each candidate function $f$. We describe algorithms that find approximate solutions to problem \ref{eq:opt} in section \ref{sec:fitting}.

$\mu_0(\tilde x)$ and $\mu_1(\tilde x)$ are not meant to be estimates of the true conditional mean functions of the outcome under the true unobserved distribution $P(X,W,Y)$. Rather, $\mu_0(\tilde x)$ and $\mu_1(\tilde x)$ are the most likely conditional means under the assumption that the true unknown effect is the synthetic effect $\tilde \tau$ and that $P(X,W) = P(X^*, W^*)$. Again, if we could confidently estimate the true conditional means, we would not need to compare different methods. In particular, because of confounding, we are concerned about our ability to accurately estimate the true conditional means in areas of low covariate support. 

As with setting the synthetic effect, estimating $\mu_0(\tilde x)$ and $\mu_1(\tilde x)$ requires making modeling choices, the most important of which is the model space $\mathcal{F}$. If a causal inference method uses conditional mean modeling over the same model space $\mathcal{F}$ to estimate the treatment effect, it will be correctly specified relative to the generative distributions $P(\tilde X, \tilde W, \tilde Y)$. This could mean that synth-validation would be biased towards selecting that method. However, we show that this bias is not observed in practice when using a space of flexible models and discuss why that may be so in section \ref{sec:disc}.

With $\mu_0(\tilde x)$, $\mu_1(\tilde x)$, and $P(R^*)$ in hand, we sample from $P(\tilde X, \tilde W, \tilde Y)$:

\begin{equation}
\begin{array}{rcl}
(\tilde x_i, \tilde w_i) & \overset{\text{I.I.D.}}{\sim} & P(X^*, W^*) \\
r_i & \overset{\text{I.I.D.}}{\sim} & P(R^*) \\
\tilde y_i & = & I_0(\tilde w_i)\mu_0(\tilde x_i) + I_1(\tilde w_i)\mu_1(\tilde x_i)+ r_i
\end{array}
\end{equation}

We do this $n$ times to get a synthetic dataset $(\tilde{\mathbf{x}}, \tilde{\mathbf{w}}, \tilde{\mathbf{y}})$. This is a semi-parametric bootstrap where the ``parametric model'' for $\tilde y$ is $I_0(\tilde w_i)\mu_0(\tilde x_i) + I_1(\tilde w_i)\mu_1(\tilde x_i)$.

We now have all of the pieces in place to be able to create and sample from generative distributions that have known average treatment effects and are informed by the observed data:

\begin{enumerate}
\item Pick a set of synthetic effects $\{\tilde \tau_1, \tilde \tau_2 \dots\}$ (section \ref{sec:synth-effect})
\item For each synthetic effect, estimate $\mu_0(\tilde x)$ and $\mu_1(\tilde x)$ (section \ref{sec:fitting})
\item For each synthetic effect, combine $\mu_0(\tilde x)$ and $\mu_1(\tilde x)$ with $P(X^*, W^*)$ in a semi-parametric bootstrap to sample from $P(\tilde X, \tilde W, \tilde Y)$
\end{enumerate}


\subsubsection{Choosing the synthetic effects}
\label{sec:synth-effect}

Setting the synthetic effect allows us to benchmark the performance of causal inference methods on the resulting synthetic data. To most closely model reality, we would want to set the synthetic effect to the true effect, but if we knew the true effect we would not need to model anything. Since we do not have the true effect, we could use plug-in estimates of it. However, as previously stated, there are many causal inference methods, each of which produces a different estimate of the causal effect, and we do not a-priori know which of them to trust. This is the problem that motivates synth-validation. 

Since we do not know the true effect, we create generative distributions that span several plausible effects. To define the set of synthetic effects, we run each causal inference method $t \in \mathcal{T}$ on the observed data and record each effect estimate $\hat\tau_t$. We set the largest and smallest plausible effects to be the median of the estimates plus or minus a number $\gamma$ times the range of all the estimates. From this span, we evenly sample $Q$ synthetic treatment effects and call these $\tilde\tau_1, \tilde\tau_2, \dots \tilde\tau_Q$. Although data-driven and theoretically motivated, this is a heuristic and we investigate the impact of the choice of $\gamma$ and $Q$ in our evaluation. 

\subsubsection{Fitting constrained conditional mean models}
\label{sec:fitting}

To create the conditional distribution $P(\tilde Y| \tilde X, \tilde W)$ we must estimate the conditional mean functions $\mu_0(\tilde x)$ and $\mu_1(\tilde x)$ by solving problem \ref{eq:opt}, which we restate here for convenience:

\begin{equation}
\mu_0(\tilde x), \mu_1(\tilde x) = 
\begin{cases}
\underset{f_0, f_1 \in \mathcal{F}}{\text{argmin}} \ \  \sum_{\mathcal{S}_0}  l(y_i, f_0(x_i)) + \sum_{\mathcal{S}_1}  l(y_i, f_1(x_i)) \\
\text{subject to:   } \frac{1}{n}\sum_i[f_1(x_i) - f_0(x_i)] = \tilde \tau
\end{cases}
\end{equation}

We present two algorithms that we use for this purpose: fit-plus-constant and constrained gradient boosting. Fit-plus-constant is simple and uses off-the shelf software, but finds minima that are suboptimal compared to constrained gradient boosting. 

\paragraph{Fit-plus-constant algorithms}
We present a simple algorithm that sub-optimally minimizes the objective while satisfying the constraint: First we use any regression method (e.g. linear regression, gradient tree boosting) with a model space $\mathcal{F}$ to fit two functions $h_0(x)$ and $h_1(x)$ to the untreated and treated subsamples of the data, respectively. These two functions represent the two potential outcomes ``surfaces''. We then optimize constants to add to each function so that the synthetic treatment effect is satisfied:

\begin{equation}
c^\dagger_0, c^\dagger_1 =
\begin{cases}
\underset{c_0, c_1}{\text{argmin}} \ \  
\begin{array}{l}
\sum_{i \in \mathcal{S}_0}  l(y_i, h_0(x_i) + c_0) + \\
\sum_{i \in \mathcal{S}_1 } l(y_i, h_1(x_i) + c_1) 
\end{array} \\
\text{subject to:   } \tilde{\tau} = \frac{1}{n}\sum_i^n \big[(h_1(x_i) + c_1) - (h_0(x_i) + c_0)\big]
\label{eq:opt-proj-consts}
\end{cases}
\end{equation}

Since $h_w(x_i)$ are pre-computed, they are constant quantities in optimization problem \ref{eq:opt-proj-consts}. Using a squared-error loss, the objective reduces to $\sum_i^n ( c_{w_i}^2 + c_{w_i}r_i)$ where $r_i= y_i - h_{w_i}(x_i,w_i)$ are the residuals of the model fits. The constraint reduces to $c_1-c_0 = \tilde{\tau} - \hat{\tau}_h$ where $\hat{\tau}_h = \frac{1}{n}\sum_i^n \big[h_1(x_i) - h_0(x_i)\big]$. This is a two-variable quadratic program with a linear constraint that is quickly solved by off-the-shelf software.

After solving for $c^\dagger_0$ and $c^\dagger_1$, we set $\mu_0(\tilde x ) = h_0(\tilde x) + c^\dagger_0$ and $\mu_1(\tilde x) = h_1(\tilde x) + c^\dagger_1$. This two-step approach (1. fit, 2. add constant) can be used with any off-the-shelf machine learning and optimization software. It is also easy to fit models with different synthetic effects because the potential outcome models $h_w$ do not need to be refit. However, for most practical model spaces $\mathcal{F}$, this algorithm is not a principled approach to finding an approximate optimum for problem \ref{eq:opt}.
 
\paragraph{Constrained gradient boosting}
We develop a better way to approximately solve problem \ref{eq:opt} using gradient boosting \cite{Friedman:2001ue}. We fit the model and satisfy the constraint in a single algorithm. Our approach is related to gradient projection methods \cite{Rosen:1961jl}.

We posit that $\mu_0(\tilde x)$ and $\mu_1(\tilde x)$ are linear combinations of $m$ basis functions $\mu_{w_m}(\tilde x) = \sum_j^m \nu_{w_j} b_{w_j}(\tilde x)$ which we will greedily learn in stages from the data. For simplicity of exposition, we use squared-error loss: $l(y,f) = (y-f)^2$.

We begin by setting each of $\nu_{0_1} = \nu_{1_1} = 1$ and setting each of the first pair of basis functions as constants so that the synthetic treatment effect is satisfied:

\begin{equation}
b_{1_1}, b_{0_1} =
\begin{cases}
\underset{c_0, c_1}{\text{argmin}} \ \ 
\begin{array}{l}
\sum_{i \in \mathcal{S}_0}  l(y_i, c_0) + \\
\sum_{i \in \mathcal{S}_1}  l(y_i, c_1) 
\end{array}\\
\text{subject to:   } \tilde{\tau} = \frac{1}{n}\sum_i^n \big[c_1 - c_0\big]
\end{cases}
\label{eq:first-opt}
\end{equation}

This is a two-variable linearly constrained quadratic program that is quickly solved with off-the-shelf software.

In each successive stage $m>1$, we use a learning algorithm with model space $\mathcal{B}$ (a ``weak learner'') to independently fit each of a pair of basis functions to the residuals of the previous fits: 

\begin{equation}
\begin{array}{rcl}
b_{0_m}(x) & = & \underset{b \in \mathcal{B}}{\text{argmin}} \sum_{i \in \mathcal{S}_0} l \left(r_{m-1_i} , b(x_i) \right) \\
b_{1_m}(x) & = & \underset{b \in \mathcal{B}}{\text{argmin}} \sum_{i \in \mathcal{S}_1} l \left(r_{m-1_i} , b(x_i) \right) 
\end{array}
\end{equation}

where $r_{m-1_i} =y_i - I_0(w_i)\mu_{0_{m-1}}(x_i) - I_1(w_i)\mu_{1_{m-1}}(x_i)$ is the residual of the fit at stage $m-1$. The most popular form of boosting uses regression trees for the model space $\mathcal{B}$, and we follow suit in our own implementation so that each basis function $b_{w_j}(x)$ is the output of a regression tree fit to the previous residual.


Presuming that the synthetic effect constraint (equation \ref{eq:constraint}) is satisfied at stage $m-1$, adding the basis functions $b_{0_m}(x)$ and $b_{1_m}(x)$ to the model breaks the constraint satisfaction because the basis functions are fit without regard to the constraint. To maintain the constraint satisfaction at stage $m$ we must set the multipliers $\nu_{0_m}$ and $\nu_{1_m}$ such that $\sum_i^n \big[ \nu_{1_m}b_{1_m}(x_i) - \nu_{0_m}b_{0_m}(x_i) \big] = 0$. We therefore solve the following optimization problem:

\begin{equation}
\nu_{0_m}, \nu_{1_m} =
\begin{cases}
\underset{c_0, c_1}{\text{argmin}} \ \ 
\begin{array}{l} 
\sum_{i \in \mathcal{S}_0} l \left( y_i, \mu_{0_{m-1}}(x_i) + c_{0}b_{0_{i_m}}(x_i) \right) + 
\lambda c_0^2 + \\
\sum_{i \in \mathcal{S}_1} l \left( y_i, \mu_{1_{m-1}}(x_i) + c_{1}b_{1_{i_m}}(x_i) \right) + 
\lambda c_1^2
\end{array} \\
\text{subject to:   } \sum_i^n \big[ c_1 b_{1_m}(x_i) - c_0 b_{0_m}(x_i) \big] = 0
\end{cases}
\label{eq:stage-opt}
\end{equation}

Since the constraint is satisfied for $m=1$ by solving problem \ref{eq:first-opt}, we have by induction that the constraint will be satisfied for any stage $m$.

Solving problem \ref{eq:stage-opt} preserves as much of the fit to the data as is possible while ensuring that the model at stage $m$ will continue to satisfy the synthetic treatment effect. This a constrained plane search that is the analogue of the unconstrained line search that takes place in standard boosting algorithms \cite{Hastie:2009fg}. The regularization terms $\lambda c_0^2$ and  $\lambda c_1^2$ take the place of the scaling parameter in standard boosting algorithms, which limits the contribution that each successive basis function has on the final fit. We discuss later why we use regularization in the optimization instead of scaling after optimization. 

Since we use trees as the basis functions in our implementation, we follow the example of standard tree boosting algorithms and directly optimize the fitted values in the leaf nodes instead of using the same multiplier across the entire tree \cite{Hastie:2009fg}. In other words, the tree-fitting step only determines the structure of each tree, not the predicted values in the leaves. Regardless, under squared-error loss, problem \ref{eq:stage-opt} is a quadratic problem with a linear constraint. The number of optimization variables is always small: either two in the case described here or two times the number of nodes allowed per tree (usually between $2^2$ to $2^5$ ) if optimizing the values in each leaf. These problems are solved very efficiently by off-the-shelf software.

The algorithm alternates between fitting a pair of basis functions and solving a small quadratic optimization problem until a user-set maximum number of stages have been fit. We use cross-validation for model selection among the parameters $m$, $\lambda$, and the tree depth. There is an important nuance when cross-validating in a constrained model space: when performing the optimizations in each stage (problem \ref{eq:stage-opt}), the constraint is evaluated over the whole dataset, even though the loss is only evaluated over the training data. The constraint does not give any information about the real outcomes in the held-out data, so using the held-out data in the constraint is still honest.

\subsection{Benchmarking causal inference methods on synthetic data}
\label{sec:run-meth}

We now have all the machinery that is necessary to create generative distributions that resemble the true unknown distribution. We now sample from these distributions and use that synthetic data, along with the known synthetic treatment effects, as benchmarks on which to evaluate our causal inference algorithms.

We take $K$ bootstrap samples from each of our $Q$ synthetic generative distributions, yielding $Q \times K$ synthetic datasets. We write each of these datasets as $d_{q,k}$ In this notation, $q$ indexes the synthetic treatment effect $\tilde\tau_q$ and its corresponding model and $k$ indexes the bootstrap sample.

We run each of our causal inference methods $t \in \mathcal{T}$ on each synthetic dataset $d_{q,k}$. This produces $|\mathcal{T}| \times Q \times K$ estimates of the synthetic treatment effect, which we call $\hat{\tilde\tau}_{t,q,k}$. 

To select a causal inference method, we first calculate the errors that each causal inference method makes in estimating the synthetic effects: $\tilde{e}_{t,q,k} = |\tilde\tau_q - \hat{\tilde\tau}_{t,q,k}|$. We select the causal inference method that has the least error averaged over all synthetic effects and bootstrap samples:

\begin{equation}
t^{\dagger} = \underset{t \in \mathcal{T}}{\text{argmin}} \sum_{q,k}^{Q,K} \tilde{e}_{t,q,k}
\label{eq:best-meth}
\end{equation}

\section{Evaluation}
\label{sec:eval}

\subsection{Experiments}

We use the set of simulations from \citet{Powers:2017wd} to test various permutations of synth-validation. The simulations are comprised of sixteen different combinations (scenarios) of treatment- and outcome-generating functions. The true average treatment effect in each of these scenarios is standardized to be zero, although each scenario has different levels of effect heterogeneity. The first eight scenarios have randomized treatment assignment (no confounding). The remaining eight scenarios are the same as the first eight, but with biased treatment assignment where subjects most likely to benefit from the treatment are more likely to receive it. We generate ten datasets (repetitions) from each scenario for a total of 160 datasets to test our algorithms on. 

For each dataset, we estimate the treatment effect using five causal inference methods. The first is a baseline method: the difference in mean outcomes between the treatment groups (raw). We also use three popular causal inference methods: linear regression of the outcome on the treatment adjusted for the covariates (adjusted), linear regression of the outcome on the treatment using a 1:1 propensity-matched sample (1:1 matched), and linear regression of the outcome on the treated weighted by the inverse of the propensity score (IPTW). Finally, we test a machine learning approach: taking the mean difference between the potential outcomes predicted by tree boosting models fit on the treated and untreated samples (boosting) \cite{Austin:2012cy}. In all cases, the methods are set up so that the estimand is the average treatment effect (ATE). All propensity scores are calculated using logistic regression of the treatment on all covariates. Matching is done on the logit of the propensity score, with calipers equal to $0.2$ times the standard deviation of the propensity score \cite{Austin:2011dc}. Weighted estimates are calculated by trimming propensities greater than $0.99$ or less than $0.01$ and stabilizing the weights \cite{Sturmer:2014kr}. We use cross-validation to select among boosting models. 

We assess the true estimation error of each method by taking the difference of the estimate and the true effect $e_t = \hat{\tau}_t - \tau$. We calculat the error from using synth-validation as $e_{t^\dagger}$ where $t^\dagger$ is the causal inference method chosen by the synth-validation (equation \ref{eq:best-meth}). We also compare synth-valdiation against an oracle selection algorithm that achieves the lowest possible error of any method selection algorithm. We calculate the error from using the oracle selector as $e_{oracle} = e_{t^{\dagger \dagger}} = \underset{t \in \mathcal{T}}{\text{min}} \ e_t$. The oracle selector is not usable outside of simulations because it uses the true errors $e_t$ to select the causal inference method, which requires knowing the true effect $\tau$ a-priori.

We test various permutations of synth-validation. Each permutation is a combination of an algorithm for fitting the conditional mean models (section \ref{sec:fitting}) and an approach for choosing the synthetic effects (section \ref{sec:synth-effect}).

\subsubsection{Fitting}
\label{sec:fitting-eval}
For fitting, we test fit-plus-constant using linear regression (FPC-linear), fit-plus-constant using gradient boosted trees (FPC-treeboost), and constrained gradient boosted trees (CGB-tree).

\subsubsection{Synthetic Effect Choice}
\label{sec:synth-effect-eval}
To choose the synthetic effects, we use the approach described in section \ref{sec:synth-effect} with $Q=5$ synthetic effects spaced out over a narrow span using $\gamma=2$ (narrow) or a wide span using $\gamma=5$ (wide). We also test using $Q=10$ synthetic effects that are the union of those in the wide and narrow spans (combined). 


\paragraph{} We test every combination of these approaches for fitting and setting the synthetic effects. In each case we use $K=20$ bootstrap samples from each resulting generative distribution (section \ref{sec:run-meth}).

\section{Results}
\label{sec:results}

\begin{figure}[h!]
\centering
\includegraphics[width=\textwidth]{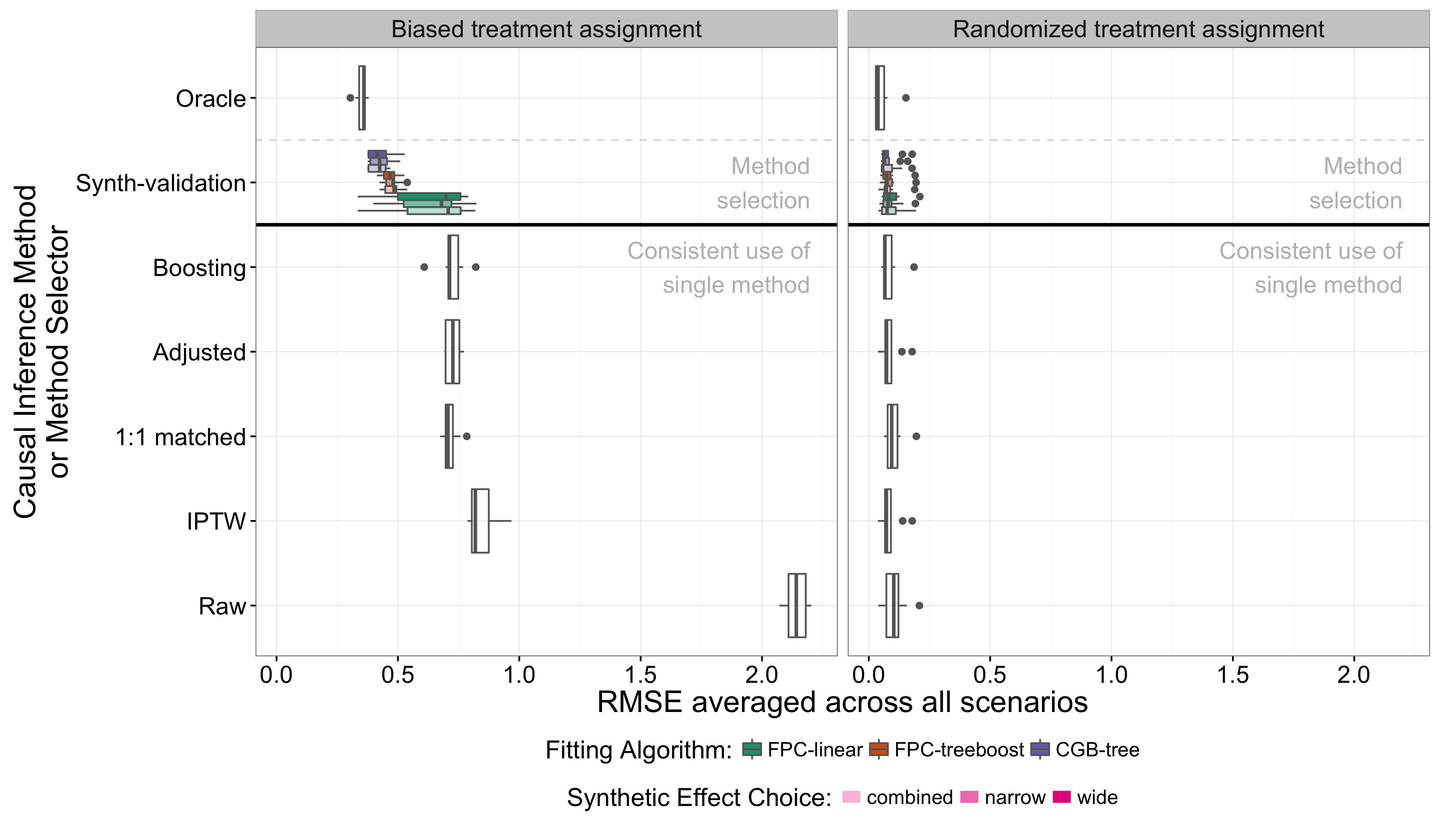} 
\caption{Average error across all scenarios in estimating the true treatment effect using each causal inference method across all scenarios or using synth-validation or an oracle to select which method to use in each scenario. Lower error is better. The left plot includes only scenarios with biased treatment assignment, the right includes only those with randomized assignment. Boxplots represent variation across the ten repetitions of the entire evaluation. We evaluate several variants of synth-validation: colors represent different fitting algorithms (sections \ref{sec:fitting-eval} and \ref{sec:fitting}) while different saturations (darkness of the color) represent different heuristics used to choose the synthetic effects (sections  \ref{sec:synth-effect-eval} and \ref{sec:synth-effect}).}
\label{fig:result}
\end{figure}

\paragraph{Performance} Our results show that all of the synth-validation variants we propose outperform the individual use of each of the causal inference methods that synth-validation selects from (figure \ref{fig:result}). Synth-validation using CBG-tree has the best performance, followed closely by synth-validation with FPC-treeboost,
and trailed by synth-validation with FPC-linear
only marginally beating the individual use of any of the causal inference methods
Synth-validation appears robust to the choice of $\gamma$ and $Q$ in our heuristic for choosing the synthetic treatment effects. No settings (narrow, wide, or combined) result in notably better performance.
The individual causal inference methods all perform on par with each other. The baseline method (raw) performs poorly in comparison to other methods.

These results are driven by performance on scenarios with biased treatment assignment. All methods work about equally well when assignment is randomized and the average effect is easy to estimate.

\begin{figure}[h!]
\centering
\includegraphics[width=\textwidth]{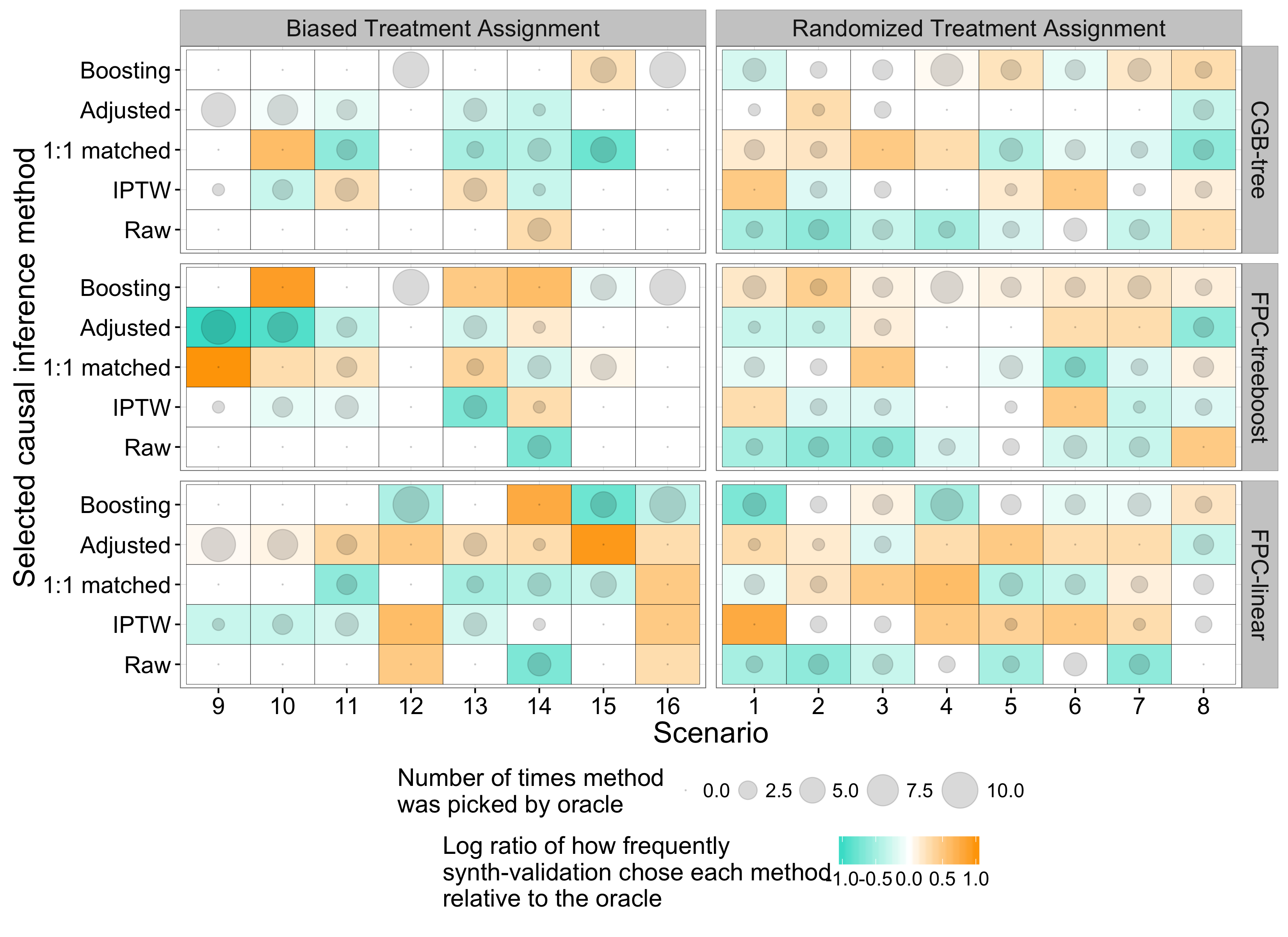} 
\caption{
Number of times that synth-validation with different fitting algorithms selected each causal inference method across the ten repetitions of each scenario, relative to the number of times each method was selected by the oracle across the same repetitions. The left panels are scenarios with biased treatment assignment, the right are scenarios with randomized treatment assignment. The top, middle, and bottom panels are the different fitting algorithms used in synth-validation (sections \ref{sec:fitting-eval} and \ref{sec:fitting}). Colors represent the number of times that synth-validation selected each method in each scenario relative to the number of times that method was selected by the oracle in that scenario (a ratio of frequencies). Orange means that synth-validation selected that method in that scenario more often than the oracle did across repetitions of each scenario; blue means that synth-validation selected that method in that scenario fewer times than the oracle did. Lighter colors (closer to white) indicate close agreement with the oracle and thus optimal performance. The size of the circles represents the absolute number of times that the oracle chose each method in each scenario. The pattern of circles is the same in the top, middle, and bottom panels since each panel compares synth-validation using a different fitting algorithm to the same comparator (the oracle).
}
\label{fig:selection}
\end{figure}

\paragraph{Algorithm biases} The oracle chooses different methods with different frequencies in different scenarios (figure \ref{fig:selection}). There are some scenarios where one method dominates (e.g. Boosting in scenario 12), but more frequently there is some variance across repetitions of the evaluation in which is the best method. The distribution of choices with synth-validation using CBG-tree most closely tracks the distribution of the oracle's choices (i.e. their checkerboard patterns are a close match). Synth-validation with FPC-linear seems to have an affinity to the adjusted method of causal inference in the scenarios with biased assignment, whereas synth-validation with FPC-treeboost seems to be biased towards boosting in scenarios with randomized assignment
\section{Discussion}
\label{sec:disc}

Our results demonstrate that synth-validation successfully lowers estimation error by choosing a causal inference method that is appropriate for the observed data.

\subsection{Components of Synth-Validation}

\subsubsection{Fitting}
CGB-tree, the most sophisticated of our fitting approaches, is the best performing. FPC-treeboost also performs well, but since the computational difference between the two is the cost of solving $m$ small quadratic programs (regardless of $n$), the two scale equally well and there is not a good argument to be made for FPC-treeboost besides that it can be implemented with off-the-shelf software.

The difference between our three fitting approaches is how ``similar'' we require the resulting synthetic outcomes to be to the original outcomes, where similarity is measured in terms of the loss in equation \ref{eq:opt}. Because of the more prudent optimization scheme, the minimum value of the loss obtained via CGB-tree should be less than that obtained by FPC-treeboost. There is no general relationship between the minimum loss of FPC-treeboost and FPC-linear, but the only difference between the two is the off-the-shelf learner (boosting vs. linear regression) used to fit $h_0$ and $h_1$. Boosting is generally considered to achieve better fits than linear models on most data, and so we generally expect FPC-treeboost to achieve a lower loss than FPC-linear. 

Based on this theory and our results, we posit that the difference in performance between algorithms that use these different fitting approaches is due to the extent to which they can produce synthetic data that ``looks like'' the real data. In practice, users are free to use cross-validation as we implement it to select not just between parameter settings of CGB-tree, but between different fitting approaches altogether. If our hypothesis holds, then algorithms using the fitting approach that yields the lowest expected MSEs while satisfying the same set of synthetic treatment effects should generally select the causal inference method that has the lowest estimation error on the real data.

\begin{figure}[h!]
\centering
\includegraphics[width=\textwidth]{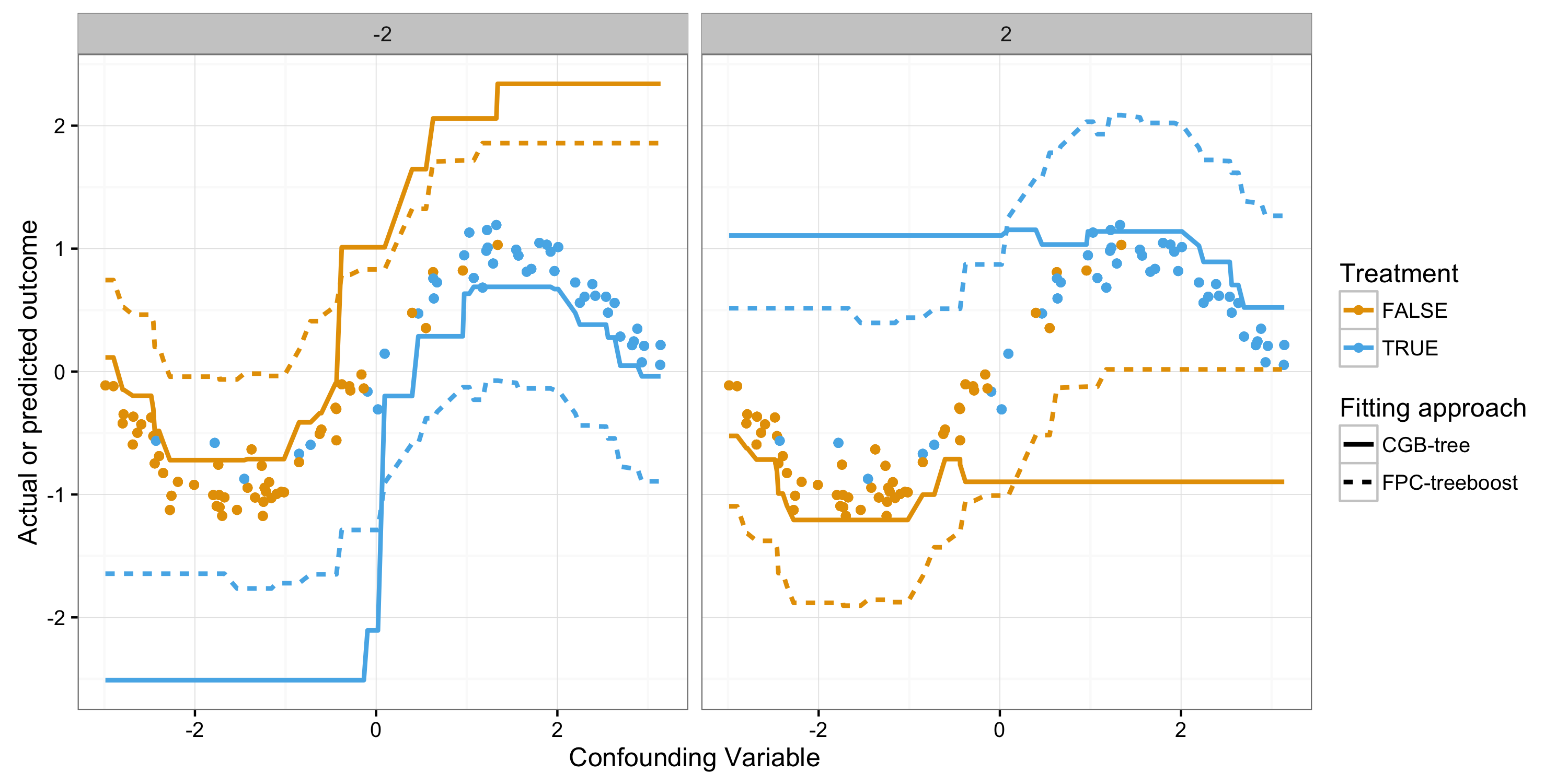} 
\caption{Illustration the behavior of different fitting approaches using a toy dataset. Lines represent the conditional mean functions estimated using CGB-tree and FPC-treeboost and points represent the original dataset. The toy dataset has a single confounder $X \sim \text{Uniform}(-\pi, \pi)$, biased treatment assignment $W \sim \text{Binom}((1+e^{-2X})^{-1})$ and an outcome that does not depend on the treatment $Y\sim \text{Normal}(\sin(X), 0.15)$. The synthetic effect is set to either $\tilde\tau=-2$ (underestimated) or $\tilde\tau=2$ (overestimated). Note that CGB-tree stays closer to the data in regions of good support (e.g. $x<0$ for the untreated), but diverges from the data in regions of poor support (e.g. $x<0$ for the untreated).}
\label{fig:didactic}
\end{figure}

The relative differences between using synth-validation vs. single causal inference methods disappear when the treatment is randomized. Primarily this is because it becomes easy to estimate average treatment effects in randomized settings and so all causal inference methods perform reasonably well (coming close to matching the performance of the oracle selector). However, even if this were not the case (e.g. if one of the causal inference methods were a strawman that always returned $\hat\tau=10$), we suspect that each fitting approach in our algorithms would perform equally well given randomized data. On nonrandomized data, CGB-tree preferentially ``moves'' potential outcomes that are in regions of low observed support for their treatment in order to satisfy the constraint while better minimizing the objective. In other words, the residuals for subjects in regions of low covariate support (given their treatment) are higher. FPC-treeboost uses a single constant to shift its initial predictions, and thus we expect the residuals of its final predictions to all be of similar magnitude (figure \ref{fig:didactic}). In randomized data, there are no regions of low support for either treatment, and we expect the predictions from CGB-tree to be very similar to those of FPC-treeboost, and consequently their downstream results should be similar as well. 

Figure \ref{fig:didactic} also illustrates the differences between synthetic datasets generated by synth-validation using CGB-tree and FPC-treeboost. Using CGB-tree, the synthesized outcomes are very close to the observed outcomes for the majority of subjects, but for a minority of them (those in regions of low support), the synthesized outcomes may be very different than the observed outcomes. Using FCP algorithms, the outcomes for all subjects are perturbed by a moderate amount. 

It is interesting that CGB-tree is not demonstrably biased towards picking any particular causal inference method. This can be seen in figure \ref{fig:selection}, where biases are visible as horizontal bands of blue or orange. FPC-linear has a clear bias for covariate-adjusted linear regression ("adjusted") and FPC-treebost seems to be biased towards boosting (visible in the right panel). Our hypothesis before performing the evaluation was that each fitting approach would favor the causal inference method with the correct outcome model specification: linear regression for FPC-linear and boosting for FPC-treeboost and CGB-tree. Our results, however, tell a more nuanced story: there are many scenarios in which boosting is not chosen as the causal inference method by CGB-tree or FPC-treeboost, or both. It may be the case that other causal inference methods often perform better on synthetic data generated by boosting models because the final target of inference is the average treatment effect, not the conditional mean of the outcome-generating function per-se. Boosting relies on estimates of unobserved potential outcomes for subjects that are in regions of low support. Even if the model is well-specified, there will be higher variance in the predictions in those regions, which could impact error in treatment effect estimation. This is especially relevant for CGB-tree, where synthetic outcomes for subjects in regions of low support are more likely to vary unexpectedly (figure \ref{fig:didactic}). This, combined with the fact that regions of low support will be less sampled from during bootstrapping, means that boosting is likely to critically misestimate potential outcomes in those regions. This helps explain why synth validation with CGB-tree appears immune to bias from correctly-specified outcome models.

\subsubsection{Synthetic Effect Choice}
The choice of the synthetic effects does not play a large role in the performance of synth-validation. Considering how close synth-validation with CGB-tree gets to the oracle in terms of performance, it seems that our heuristic works well enough to advocate its use. There are other heuristics that may also work. Considering that changing $\gamma=2$ to $\gamma=5$ in our heuristic did not change performance, using other heuristics may not have a large impact either. In practice the choice of heuristic or of $\gamma$ should be based on expert knowledge of the domain: if few previous studies exist, it would be wiser to use a larger value of $\gamma$. 


It is likely that more misspecification of the synthetic treatment effects would degrade performance. However, employing our heuristic, misspecification of the synthetic effects only occurs if no causal inference method comes close to correctly estimating the effect. In such a situation, selection of the best method is largely a moot point: the study is bound to have large bias regardless.

\subsection{Future Work}

\paragraph{Further evaluation} We have focused here on the conceptual and methodological advancements necessary to perform dataset-specific selection of causal inference methods. Further simulation studies using larger sets of scenarios and causal inference methods will be helpful in teasing out and explaining differences in performance. It may also be possible to perform an equivalent evaluation on a set of real observational datasets where the true treatment effect is known to a high degree of confidence from large randomized experiments (e.g. \citet{Madigan2014}). 

\paragraph{Extension to other classes of outcomes} For the sake of brevity, we have limited the current work to algorithms that are formulated for real-valued outcomes and squared-error loss. Alternative losses are easily accommodated: as long as they are convex, the resulting optimization problems (equations \ref{eq:opt-proj-consts}, \ref{eq:first-opt}, and \ref{eq:stage-opt}) will still be convex, if not quadratic. However, many outcomes of interest are binary or time-to-event, which requires more than changing the loss. We have generalized our approach to other kinds of outcomes using link functions (e.g. for binary outcomes, treatment effects are risk differences and we model the log-odds of the outcome in a continuous space). This creates nonlinear constraints, but since each optimization problem is still small, the algorithms remain tractable. In fact, these nonlinear constraints are what motivate the use of a regularizer in equation \ref{eq:stage-opt} instead of a scaling factor: scaling the solution to a nonlinearly constrained optimization problem takes it off of the constraint surface. We reserve further discussion and evaluation of these extensions for a later date.

\paragraph{Impact of unmeasured confounding} No modeling approach can capture the effects of real unobserved confounding. However, some unobserved confounding can be simulated within synth-validation by removing certain covariates from the synthetic datasets. Variable importance metrics for each covariate in the fit models could be used to choose which covariates to remove; removing covariates with high variable importance would simulate high unobserved confounding and vice-versa. However, there remains the possibility that the unobserved confounders have relationships to the outcome that are fundamentally different from those between the measured covariates and the outcome, in which case even this kind of artificial censoring would not result in synthetic datasets that are good proxies for the original. Further study will be necessary to elucidate the impact of different kinds of unmeasured confounding.

\paragraph{Software} We plan to release user-friendly open-source software packages that implement our algorithms for causal inference method selection.

\section{Conclusion}

To our knowledge, we present the first formal approach to select causal inference methods in a way that is tailored to a given dataset. We show that the best-performing causal inference method for one dataset is not necessarily the best for another. We present synth-validation, a procedures that estimates how well causal inference methods will estimate an average treatment effect in the context of a given dataset. We evaluate synth-validation using a large number of diverse simulated datasets with known treatment effects. Using synth-validation results in a meaningful and significant decrease in the expected error of estimating the average treatment effect relative to the consistent use of any single causal inference method. We suggest that practitioners of observational studies in healthcare, business, and other policy domains use synth-validation to improve treatment effect estimation and contribute to better decision-making.


\bibliographystyle{plainnat}
\bibliography{bibliography}

\begin{thebibliography}{27}
\providecommand{\natexlab}[1]{#1}
\providecommand{\url}[1]{\texttt{#1}}
\expandafter\ifx\csname urlstyle\endcsname\relax
  \providecommand{\doi}[1]{doi: #1}\else
  \providecommand{\doi}{doi: \begingroup \urlstyle{rm}\Url}\fi

\bibitem[Antonelli et~al.(2016)Antonelli, Cefalu, Palmer, and
  Agniel]{Antonelli:2016ve}
Joseph Antonelli, Matthew Cefalu, Nathan Palmer, and Denis Agniel.
\newblock {Double robust matching estimators for high dimensional confounding
  adjustment}.
\newblock \emph{arXiv.org}, December 2016.

\bibitem[Austin(2011)]{Austin:2011dc}
Peter~C Austin.
\newblock {Optimal caliper widths for propensity-score matching when estimating
  differences in means and differences in proportions in observational
  studies}.
\newblock \emph{Pharmaceutical Statistics}, 10\penalty0 (2):\penalty0 150--161,
  March 2011.

\bibitem[Austin(2012)]{Austin:2012cy}
Peter~C Austin.
\newblock {Using Ensemble-Based Methods for Directly Estimating Causal Effects:
  An Investigation of Tree-Based G-Computation}.
\newblock \emph{Multivariate Behavioral Research}, 47\penalty0 (1):\penalty0
  115--135, February 2012.

\bibitem[Colson et~al.(2016)Colson, Rudolph, Zimmerman, Goin, Stuart, van~der
  Laan, and Ahern]{Colson:2016fu}
K~Ellicott Colson, Kara~E Rudolph, Scott~C Zimmerman, Dana~E Goin, Elizabeth~A
  Stuart, Mark van~der Laan, and Jennifer Ahern.
\newblock {Optimizing matching and analysis combinations for estimating causal
  effects}.
\newblock \emph{Scientific reports}, pages 1--11, March 2016.

\bibitem[Efron and Tibshirani(1993)]{Efron:1993dc}
Bradley Efron and Robert~J Tibshirani.
\newblock {An Introduction to the Bootstrap}.
\newblock Springer US, Boston, MA, 1993.

\bibitem[Franklin et~al.(2014)Franklin, Rassen, Schneeweiss, and
  Polinski]{Franklin:2014kz}
Jessica~M Franklin, Jeremy~A Rassen, Sebastian Schneeweiss, and Jennifer~M
  Polinski.
\newblock {Plasmode simulation for the evaluation of pharmacoepidemiologic
  methods in complex healthcare databases}.
\newblock \emph{Computational Statistics {\&} Data Analysis}, 72:\penalty0
  219--226, April 2014.

\bibitem[Friedman(2001)]{Friedman:2001ue}
J~H Friedman.
\newblock {Greedy function approximation: a gradient boosting machine}.
\newblock \emph{Annals of statistics}, 2001.

\bibitem[Hannan(2008)]{Hannan:2008gh}
Edward~L Hannan.
\newblock {Randomized Clinical Trials and Observational Studies}.
\newblock \emph{Jac}, 1\penalty0 (3):\penalty0 211--217, June 2008.

\bibitem[Hastie et~al.(2009)Hastie, Tibshirani, and Friedman]{Hastie:2009fg}
Trevor Hastie, Robert Tibshirani, and Jerome Friedman.
\newblock \emph{{The Elements of Statistical Learning}}.
\newblock Springer New York, New York, NY, 2009.

\bibitem[Hill(2011)]{Hill2011}
Jennifer~L Hill.
\newblock {Bayesian Nonparametric Modeling for Causal Inference}.
\newblock \emph{Journal of Computational and Graphical Statistics}, 20\penalty0
  (1):\penalty0 217--240, January 2011.

\bibitem[Iacus et~al.(2015)Iacus, King, and Porro]{Iacus}
S~M Iacus, G~King, and G~Porro.
\newblock {A Theory of Statistical Inference for Matching Methods in Applied
  Causal Research}.
\newblock \emph{URL: http://gking harvard edu {\ldots}}, 2015.

\bibitem[King(2005)]{King2005}
G~King.
\newblock {The Dangers of Extreme Counterfactuals}.
\newblock \emph{Political Analysis}, 14\penalty0 (2):\penalty0 131--159,
  November 2005.

\bibitem[Leacy and Stuart(2013)]{Leacy:2013fs}
Finbarr~P Leacy and Elizabeth~A Stuart.
\newblock {On the joint use of propensity and prognostic scores in estimation
  of the average treatment effect on the treated: a simulation study}.
\newblock \emph{Statistics in Medicine}, 33\penalty0 (20):\penalty0 3488--3508,
  October 2013.

\bibitem[Lee et~al.(2009)Lee, Stuart, and Lessler]{Lee2010}
Brian~K Lee, Elizabeth~A Stuart, and Justin Lessler.
\newblock {Improving propensity score weighting using machine learning}.
\newblock \emph{Statistics in Medicine}, 29\penalty0 (3):\penalty0 n/a--n/a,
  2009.

\bibitem[Madigan et~al.(2014)Madigan, Stang, Berlin, Schuemie, Overhage,
  Suchard, Dumouchel, Hartzema, and Ryan]{Madigan2014}
David Madigan, Paul~E Stang, Jesse~A Berlin, Martijn Schuemie, J~Marc Overhage,
  Marc~A Suchard, Bill Dumouchel, Abraham~G Hartzema, and Patrick~B Ryan.
\newblock {A Systematic Statistical Approach to Evaluating Evidence from
  Observational Studies}.
\newblock \emph{Annual Review of Statistics and Its Application}, 1\penalty0
  (1):\penalty0 11--39, January 2014.

\bibitem[Powers et~al.(2017)Powers, Qian, Jung, Schuler, Shah, Hastie, and
  Tibshirani]{Powers:2017wd}
Scott Powers, Junyang Qian, Kenneth Jung, Alejandro Schuler, Nigam~H Shah,
  Trevor Hastie, and Robert Tibshirani.
\newblock {Some methods for heterogeneous treatment effect estimation in
  high-dimensions}.
\newblock \emph{arXiv.org}, June 2017.

\bibitem[Robins et~al.(1992)Robins, Mark, and Newey]{Robins1992}
James~M Robins, Steven~D Mark, and Whitney~K Newey.
\newblock {Estimating Exposure Effects by Modelling the Expectation of Exposure
  Conditional on Confounders}.
\newblock \emph{Biometrics}, 48\penalty0 (2):\penalty0 479, June 1992.

\bibitem[Rosen(1961)]{Rosen:1961jl}
J~B Rosen.
\newblock {The Gradient Projection Method for Nonlinear Programming. Part II.
  Nonlinear Constraints}.
\newblock \emph{Journal of the Society for Industrial and Applied {\ldots}},
  9\penalty0 (4):\penalty0 514--532, 1961.

\bibitem[Rubin(2010)]{Rubin2010}
Donald~B Rubin.
\newblock {On the limitations of comparative effectiveness research.}
\newblock \emph{Statistics in Medicine}, 29\penalty0 (19):\penalty0 1991--5--
  discussion 1996--7, August 2010.

\bibitem[Rubin and Rosenbaum(1983)]{ROSENBAUM1983}
Donald~B Rubin and Paul~R Rosenbaum.
\newblock {The central role of the propensity score in observational studies
  for causal effects}.
\newblock \emph{Biometrika}, 70\penalty0 (1):\penalty0 41--55, 1983.

\bibitem[Schuler and Rose(2017)]{Schuler:2017cq}
Megan~S Schuler and Sherri Rose.
\newblock {Targeted Maximum Likelihood Estimation for Causal Inference in
  Observational Studies}.
\newblock \emph{American Journal of Epidemiology}, 185\penalty0 (1):\penalty0
  65--73, January 2017.

\bibitem[Setoguchi et~al.(2008)Setoguchi, Schneeweiss, Brookhart, Glynn, and
  Cook]{Setoguchi2008}
Soko Setoguchi, Sebastian Schneeweiss, M~Alan Brookhart, Robert~J Glynn, and
  E~Francis Cook.
\newblock {Evaluating uses of data mining techniques in propensity score
  estimation: a simulation study}.
\newblock \emph{Pharmacoepidemiology and Drug Safety}, 17\penalty0
  (6):\penalty0 546--555, 2008.

\bibitem[Shmueli(2010)]{Shmueli:2010ec}
Galit Shmueli.
\newblock {To Explain or to Predict?}
\newblock \emph{Statistical Science}, 25\penalty0 (3):\penalty0 289--310,
  August 2010.

\bibitem[Shortreed and Ertefaie(2017)]{Shortreed:2017fk}
Susan~M Shortreed and Ashkan Ertefaie.
\newblock {Outcome-adaptive lasso: Variable selection for causal inference}.
\newblock \emph{Biometrics}, 163:\penalty0 1149--12, March 2017.

\bibitem[Stuart(2010)]{Stuart2010}
Elizabeth~A Stuart.
\newblock {Matching Methods for Causal Inference: A Review and a Look Forward}.
\newblock \emph{Statistical Science}, 25\penalty0 (1):\penalty0 1--21, February
  2010.

\bibitem[Stuart et~al.(2013)Stuart, DuGoff, Abrams, Salkever, and
  Steinwachs]{Stuart:2013dt}
Elizabeth~A Stuart, Eva DuGoff, Michael Abrams, David Salkever, and Donald
  Steinwachs.
\newblock {Estimating Causal Effects in Observational Studies Using Electronic
  Health Data: Challenges and (some) Solutions}.
\newblock \emph{eGEMs (Generating Evidence {\&} Methods to improve patient
  outcomes)}, 1\penalty0 (3):\penalty0 1--12, December 2013.

\bibitem[Sturmer et~al.(2014)Sturmer, Wyss, Glynn, and
  Brookhart]{Sturmer:2014kr}
T~Sturmer, R~Wyss, R~J Glynn, and M~A Brookhart.
\newblock {Propensity scores for confounder adjustment when assessing the
  effects of medical interventions using nonexperimental study designs.}
\newblock \emph{Journal of Internal Medicine}, 275\penalty0 (6):\penalty0
  570--580, June 2014.

\end{thebibliography}

\end{document}